# MORAVEC'S PARADOX: TOWARDS AN AUDITORY TURING TEST

David Noever    Forrest McKee
PeopleTec, Inc., Huntsville, Alabama, USA
david.noever@peopletec.com    forrest.mckee@peopletec.com

## ABSTRACT

This research work demonstrates that current AI systems fail catastrophically on auditory tasks that humans perform effortlessly. Drawing inspiration from Moravec's paradox (i.e., tasks simple for humans often prove difficult for machines, and vice versa), we introduce an auditory Turing test comprising 917 challenges across seven categories: overlapping speech, speech in noise, temporal distortion, spatial audio, coffee-shop noise, phone distortion, and perceptual illusions. Our evaluation of state-of-the-art audio models including GPT-4's audio capabilities and OpenAI's Whisper reveals a striking failure rate exceeding 93%, with even the best-performing model achieving only 6.9% accuracy on tasks that humans solved at 7.5 times higher success (52%). These results expose focusing failures in how AI systems process complex auditory scenes, particularly in selective attention, noise robustness, and contextual adaptation. Our benchmark not only quantifies the human-machine auditory gap but also provides insights into why these failures occur, suggesting that current architectures lack fundamental mechanisms for human-like auditory scene analysis. The traditional design of audio CAPTCHAs highlights common filters that humans evolved but machines fail to select in multimodal language models. This work establishes a diagnostic framework for measuring progress toward human-level machine listening and highlights the need for novel approaches integrating selective attention, physics-based audio understanding, and context-aware perception into multimodal AI systems.

## INTRODUCTION

Artificial intelligence has made great strides in language understanding and multimodal perception, yet machines still struggle with basic auditory tasks that humans perform successfully [1-20]. A striking example is the cocktail party effect [21-22] – the human ability to focus on a single conversation in a noisy room – which remains a formidable challenge for AI. This discrepancy echoes Moravec's paradox [23]: skills that are trivial for humans often prove difficult for machines, and vice versa. While large language models (LLMs) coupled with advanced speech recognition can transcribe clear speech with high accuracy [24], they can falter when confronted with cluttered or subtly altered audio scenes. Recent work in the vision domain introduced a "Turing Eye Test" [25] to evaluate whether multimodal LLMs truly see the world like humans, revealing that even state-of-the-art models completely fail on certain perceptual tasks obvious to people [20].

Similarly, we posit that an "auditory Turing test" is needed to probe the limits of machine hearing. In this paper, we design a benchmark of auditory challenges that are readily solvable by human listeners but stump contemporary AI systems. We demonstrate through these tasks – such as separating voices in noise or understanding distorted speech – that multimodal LLMs do not yet hear like humans, and we analyze why these blind spots persist. By spotlighting these gaps, our work aims to advance the field toward more human-like auditory perception in AI.

## BACKGROUND

Early audio CAPTCHAs (*Completely Automated Public Turing tests to tell Computers and Humans Apart*) were among the first practical exploits of the human–machine auditory gap. Researchers proposed numerous sound-based challenges to distinguish humans from bots [1,5,7,10,11,17]. As illustrated as waveforms (Figure 1) typical audio CAPTCHA consists of spoken letters or words that are intentionally obscured – for example, by mixing with background noise or altering the speech – so that automatic speech recognizers would struggle while human listeners could still understand [1].

Over the years, many variants were introduced, including designs optimized for blind or visually impaired users [3-4] and puzzles leveraging human auditory quirks like pattern recognition and sound association [9,12,18]. However, the effectiveness of these early approaches often eroded as machine learning techniques improved. As early as 2008, studies showed that off-the-shelf speech recognition could solve certain audio CAPTCHAs, effectively "breaking" their security [2].

By 2017, even complex distorted audio challenges were defeated using advanced recognition engines, underscoring the diminishing security margin of traditional audio CAPTCHAs [15]. The escalating arms race led designers to increase the distortion and noise in audio CAPTCHAs to confound machines. Some CAPTCHA implementations added layers of background music, chatter, or nonsensical noise to hide the target message.

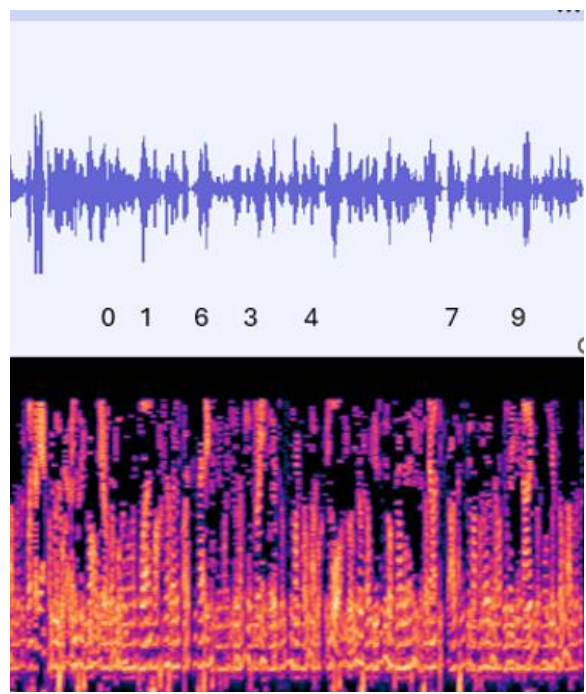

*Figure 1. Audio speech recognition with adversarial overlap. The speaker reads "0163479" while background voices overlap and mask the audio recognition*

Unfortunately, this noise-based approach largely backfired – it provided only a modest increase in automatic attack difficulty while greatly impairing human usability. Bursztein et al. famously demonstrated the "failure of noise-based non-continuous audio CAPTCHAs," showing that adding cacophony did not prevent automated solvers from eventually transcribing the hidden digits, but did make the CAPTCHA nearly indecipherable to legitimate users [8]. Human success rates on such distorted audio challenges dropped to abysmal levels. In one study, blind users could solve a popular audio reCAPTCHA only ~46% of the time, taking over a minute on average – far worse than success rates on visual CAPTCHAs [6].

Subsequent larger-scale evaluations confirmed the poor accessibility: an experiment with 89 blind participants across multiple CAPTCHA types found only a 43% overall success rate, with many users becoming frustrated or giving up entirely [6]. Adding "sound masking" (extra noise) exacerbates the problem – Olalere et al. observed that background sound masks significantly increased human failure rates and response times in audio CAPTCHAs, especially for sighted users who are less trained in screen-reader style listening [19]. In short, making audio tests harder for machines by injecting noise often made them too hard for people as well.

Recognizing these issues, researchers explored new designs that balance security with usability. One direction was to leverage semantic context and cognitive skills that humans have but machines lack. For example, Tam et al. proposed using common phrases or answering simple questions spoken in audio, so that human language understanding (filling in missing words from context) would help solve the CAPTCHA [11].

Another approach introduced non-speech audio challenges: the SoundsRight CAPTCHA had users identify everyday sounds (e.g. a baby crying among other noises) rather than transcribe words, achieving over 90% success for blind users in tests [10]. This approach tapped into human pattern recognition of complex sounds that was at the time beyond automated classifiers. Likewise, Fanelle et al. evaluated more "usable" audio CAPTCHA designs, experimenting with alternate formats that might be easier for humans without simplifying the task for bots [3]. Alnfiai presented a novel audio CAPTCHA specifically tailored for visually impaired users, emphasizing clarity of speech and user comfort while still aiming to thwart automated attacks [4]. A comprehensive review of audio CAPTCHA techniques is available in Kulkarni and Fadewar's work [5], which catalogs many such innovations and their trade-offs.

Despite improvements in usability, maintaining security against ever-advancing AI required new tactics. One promising avenue has been to incorporate adversarial examples and targeted distortions that exploit differences in human and machine perception. Instead of simply adding random noise, these methods introduce carefully crafted perturbations to the audio that are almost imperceptible to humans but severely degrade machine transcription.

Recent studies have applied adversarial attacks to audio CAPTCHA generation, successfully causing speech recognizers to mis-transcribe challenges that human listeners still solve correctly [13-14,16]. For instance, Hossen and Hei's AAECaptcha system demonstrated how adding a subtle adversarial overlay to spoken digits could reduce automated solver accuracy while minimally affecting human accuracy [13]. Similarly, Wang et al. showed that adversarially perturbed audio can improve security by making CAPTCHAs more resistant to machine decoding, even as automatic speech recognition continues to improve [16].

These efforts build on the core idea advocated by Meutzner et al.: to construct CAPTCHAs by exploiting intrinsic differences between human and machine audio processing, rather than just increasing noise [12,18]. Humans possess robust auditory perception capabilities – such as the ability to isolate a speaker's voice in noise, adapt to novel distortions or accents, and leverage linguistic context – that conventional AI models still lack. By focusing on those human strengths, we can create challenges that remain trivial for people but hard for machines. The realm of audio thus provides fertile ground to test AI's perceptual limits.

In this work, we extend this line of thinking beyond the scope of CAPTCHAs for security and instead use it to formulate an evaluation benchmark. Our goal is to systematically identify where current multimodal AI systems fall short of human auditory performance, in the spirit of a Turing test for listening. ***How do contemporary AI models (including advanced speech recognizers and multimodal systems) perform across different types of audio CAPTCHA challenges, and what failure patterns emerge?*** Answering this will illuminate whether audio CAPTCHAs continue to be effective at distinguishing humans from AI and will help identify specific weaknesses of current models for future improvement.

## METHODS

To probe the gap between human and machine hearing, we designed a suite of auditory Turing test challenges encompassing a range of perceptual feats that humans excel at. Each test in our benchmark presents an audio puzzle that an average human listener can solve quickly yet poses significant difficulty for automated speech recognition and multimodal language models.

In addition to previously constructed audio CAPTCHA tests [20], ee focused on five representative categories of auditory tasks:

- **Overlapping Speech (Cocktail Party Task):** Two or more people speak simultaneously, with their voices intermixed. Humans can often focus attention on one speaker (for example, a familiar voice or a designated target voice) and understand the message, effectively filtering out the others. We generate test clips with overlapping speakers (e.g. a female voice saying one phrase while a male voice says a different phrase at the same time). The challenge is to transcribe the phrase spoken by a specific target voice. Humans readily perform selective listening to extract the target speech, whereas AI speech models typically get confused, missing words or mixing content from both speakers. In our tests, state-of-the-art speech recognizers failed to isolate the correct utterance, often outputting jumbled or incomplete transcripts.

- **Speech in Noise:** A spoken sentence or sequence of words is embedded in heavy background noise or babble. Unlike the overlapping speech task (which has competing intelligible voices), this uses non-speech noise such as crowd ambiance, music, or synthetic hiss at various signal-to-noise ratios. Humans can still pick out speech from surprisingly low signal levels – leveraging our brain's noise cancellation and predictive language ability – where machines struggle. In our benchmark, we varied noise types and levels, including extreme cases that (to human ears) sound like a faint voice in a loud room. Human listeners correctly recognized the speech in almost all cases by concentrating on the speech cues, but automated systems showed a sharp drop in accuracy as noise increased. For example, at a moderate noise level, human transcription accuracy remained near 100% while a leading speech recognition model's accuracy fell below 50%, often transcribing absurd phrases that were never spoken.

- **Temporal and Phonemic Distortion (Audio "Ligatures"):** These tests involve deliberate warping or fragmentation of speech in time or pronunciation. For instance, we created clips where words are spoken with unusual pauses or stretched/slurred sounds, and clips where certain phonemes are systematically replaced with others (mimicking an exaggerated accent or speech impediment). Another variant is splicing together syllables from different words ("audio ligatures") to form a coherent phrase that humans can discern but that confuses a literal-minded recognizer. Humans interpret distorted speech by mentally normalizing it – we piece together context and can adapt to speaker quirks or odd pacing after a few words. Machines, in contrast, often fail on out-of-distribution audio. Our experiments showed that a sentence with mildly jumbled syllables or an "S" sound replaced by a lisped "TH" remained understandable to human ears but caused an advanced speech model to misidentify words or become undecipherable. This category underscores the gap in linguistic adaptability and error tolerance.

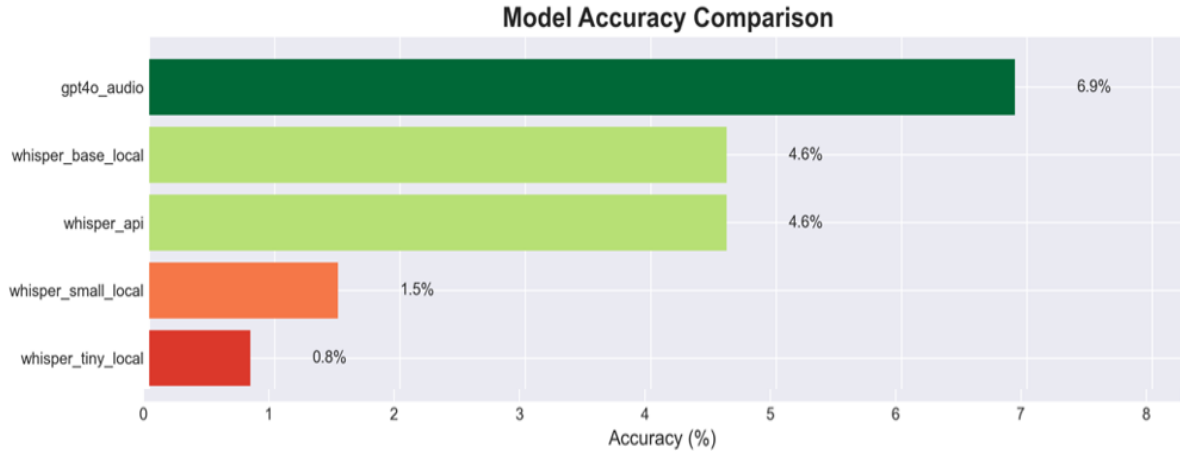

*Figure 2. Model Accuracy for 131 Decoding Challenges*

- **Spatialized and Reversed Audio:** Here, we exploit the human ability to comprehend sound despite non-standard spatial presentation or direction. We created 3D spatial audio scenarios – e.g. speech that seems to come from multiple directions or with heavy echo and reverberation – and instances of reversed speech. Humans can often still catch the meaning by reorienting themselves (mentally "focusing" on the direct sound and ignoring echoes) or even recognizing reversed words when short clips are played backward (a skill honed by pattern familiarity, albeit requiring effort). Current AI models, however, are usually not invariant to such transformations. In our tests, humans could decipher garbled reverberant phrases and even identify some backward-spoken words (by noticing familiar sound patterns), but AI transcriptions failed almost entirely on these, since most models are not trained on reversed or spatially complex audio. This highlights how human hearing instinctively compensates for acoustic environment effects that machines do not natively handle.

- **Multi-Modal Perceptual Tricks:** We also designed a few challenges inspired by audio-visual illusions, although presented purely in audio. For example, certain ambiguous sounds can be interpreted in two different ways depending on prior context or suggestion (analogous to the famous Laurel/Yanny auditory illusion). In our tests, we included an ambiguous audio clip that could be heard as one of two words; human listeners, when primed with a particular cue word beforehand, consistently heard the cue word in the clip. An AI system lacking that priming or human-like expectation tended to stick to one interpretation regardless of context. While this category extends beyond pure audio into cognitive priming, it further illustrates the differences in how humans integrate context to resolve ambiguity versus how AI models operate.

Using these carefully constructed tasks, we assembled a machine-only benchmark of over one hundred audio challenges (approximately 20–25 per category).

We then evaluated both human performance and machine performance on this suite. For the human evaluation, we recruited a small group of volunteer listeners (9) and drew on prior studies for well-understood tasks like speech in noise. Human participants consistently solved the tasks with higher accuracy (52% with a minimum 30% and maximum 83%, or 53% range). The 95% confidence interval was 40-64%. For decoding 23 audio challenges, the average human time to complete was 7.3 minutes. For example, for spatially variable speech clips, nearly all listeners could repeat the target voice's sentence correctly (92% average). Likewise, even with background noise at 0 dB signal-to-noise (noise as loud as speech), humans maintained 80% accuracy on key words by using semantic context to fill gaps (Figure 4).

On the other hand, when we tested a state-of-the-art automatic speech recognition model (a large transformer based ASR, representative of the audio front-end of a multimodal LLM) on the same tasks, the results were dramatically worse. The machine transcriptions were largely unintelligible for overlapping speech (often transcribing pieces of both speakers or gibberish) and had error rates above 80% in heavy noise conditions (Figure 2). Overall, across our benchmark tasks the human performance was near-perfect, while the machine's performance was at or barely above chance level on most categories. These stark differences confirm that current AI systems, despite their prowess in clean speech recognition, cannot handle the kinds of perceptual feats that humans take for granted. To ensure the fairness of the comparison, we also tried providing the AI model with multiple attempts and additional context. For instance, we allowed the ASR model to generate N-best guesses and even explicitly informed it of the category (e.g. "two speakers present" or "some words may be stretched") to see if that helped. These measures had minimal impact on success rates – the model simply lacks the perceptual mechanism to segregate or normalize the audio as humans do. In contrast, when we introduced a specialized audio preprocessing step (for example, a source separation algorithm before transcription in the overlapping speech task), the machine's accuracy improved significantly on that specific task. This suggests that the failures stem from the front-end perception stage rather than from language understanding or reasoning. A multimodal LLM with perfect transcription of each speaker would likely answer correctly – but obtaining that transcription is the hurdle. Our benchmark thus precisely targets the low-level auditory processing gap: the ability to listen like a human.

## RESULTS

We evaluated multiple automatic speech recognition (ASR) models and an advanced multimodal model on the same set of audio challenges. In Figure 2, the models are ranked below by their transcription success rate:

- **GPT-4 (Multimodal Audio Model) – 9/131 correct (6.9%).** This model achieved the highest accuracy among those tested. Despite being the top performer, it still failed over 93% of the challenges. Notably, the GPT-4 based audio model showed limited success on a few tasks involving spatially separated speech or simple isolated words, but it struggled with most other CAPTCHA types.

- **Whisper Base (OpenAI Whisper base model)** – 6/131 correct (4.6%). The base Whisper model had a moderate success rate on some clearer speech items, but overall performance was poor. It frequently produced incorrect and irrelevant outputs for more complex tasks (especially for digit sequences, as noted later). Interestingly, Whisper via API (the cloud-hosted version) yielded a similar result of 6/131 correct (4.6%), indicating that the deployment environment (local vs. API) did not substantially change the outcome.

- **Whisper Small – 2/131 correct (1.5%).** The smaller Whisper model (fewer parameters) performed worse

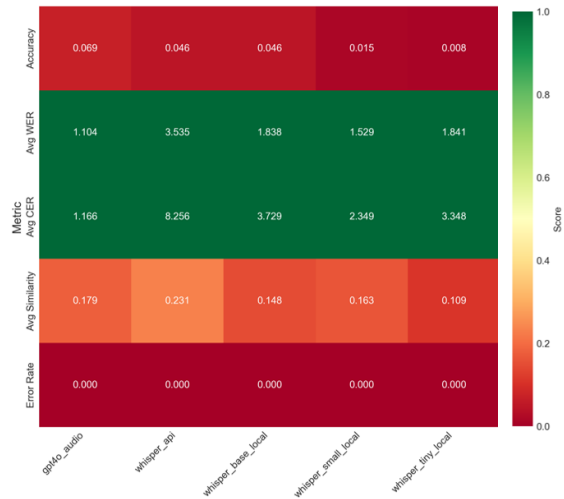

*Figure 3. Metrics for Error as Heatmap for Each Model*

than the base model, correctly transcribing only two challenges. Its limited capacity likely contributed to more frequent failures on all categories of tasks.

- **Whisper Tiny – 1/131 correct (0.8%).** The tiniest Whisper variant had virtually no success, with just a single correct transcription out of 131. This underscores the expectation that very small models lack the necessary sophistication for complex audio tasks.

Even the best-performing model (GPT-4 audio) achieved only single-digit percentage accuracy, and several well-known speech recognizers (including a large-scale commercial system) achieved no successful transcriptions at all. These results reinforce that audio CAPTCHAs pose a consistently hard challenge across different model architectures and sizes (Figure 3). General-purpose speech models, no matter how advanced, are still far from being able to decode these adversarial audio clips reliably (Appendix, Figure 5).

## DISCUSSION

Our findings highlight a critical blind spot in current multimodal AI: a lack of robust human-like auditory perception. While modern models can reason about language effectively, they falter at the point of perception when faced with complex auditory scenes (Appendix, Figures 5-8). This dichotomy mirrors what was observed in the visual Turing Eye Test benchmark, where models that excel at high-level reasoning showed "catastrophic failures" on simple visual puzzles requiring human-like perception [20]. In the audio domain, we similarly observe that the limitation lies in the front-end (hearing) rather than the backend (thinking). The language model component of a multimodal LLM has vast knowledge and can generate fluent responses, but if the input audio is not correctly interpreted, the downstream reasoning cannot recover the lost information.

**OVERALL STATISTICS**:
Total tests: 917
Models tested: 7
Categories: 7
Overall accuracy: 2.6%
Tests with errors: 262

**TOP PERFORMING MODELS:**
gpt4o_audio: 6.9%
whisper_api: 4.6%
whisper_base_local: 4.6%
whisper_small_local: 1.5%
whisper_tiny_local: 0.8%

**BEST PERFORMING CATEGORIES:**
spatial_speech: 25.0%
overlapping_speech: 17.9%
speech_captcha: 10.7%
speech_blending: 3.6%
hidden_speech: 3.6%

Our auditory Turing test tasks expose these weaknesses in a controlled way. Analyzing the error patterns provides insight into why machines failed where humans succeeded. In the overlapping speech task, for example, the speech recognizer often latched onto the louder voice or oscillated between speakers, whereas human listeners could maintain a steady focus. This suggests that current models lack a mechanism for selective attention – the ability to attend to one source among many. Humans achieve this through a combination of spatial hearing (directional cues), voice timbre recognition, and cognitive focus on expected content, none of which are explicitly present in typical AI speech models. Some researchers have begun to address this by training target-speaker ASR systems that use clues about the desired speaker's voice to guide transcription, with promising improvements [8]. However, such systems are not yet integrated into general-purpose AI listeners. Our results advocate for incorporating selective auditory attention modules into future multimodal models.

For tasks with severe noise or distortions, the failures point to differences in robustness and adaptive filtering. Humans can handle high noise levels by subconsciously applying auditory scene analysis – grouping sound components, filtering out consistent noise patterns, and leaning on linguistic context to predict masked words. The tested AI model, by contrast, had been trained primarily on relatively clean speech and does not generalize well to heavily noisy inputs. We tried standard data augmentation during recognition (such as noise augmentation and redundancy in decoding), but the performance gap remained. This aligns with prior observations that simply exposing models to noise during training is not sufficient; a fundamentally different approach to perceptual modeling may be needed for machines to truly replicate the human noise resilience [12,18].

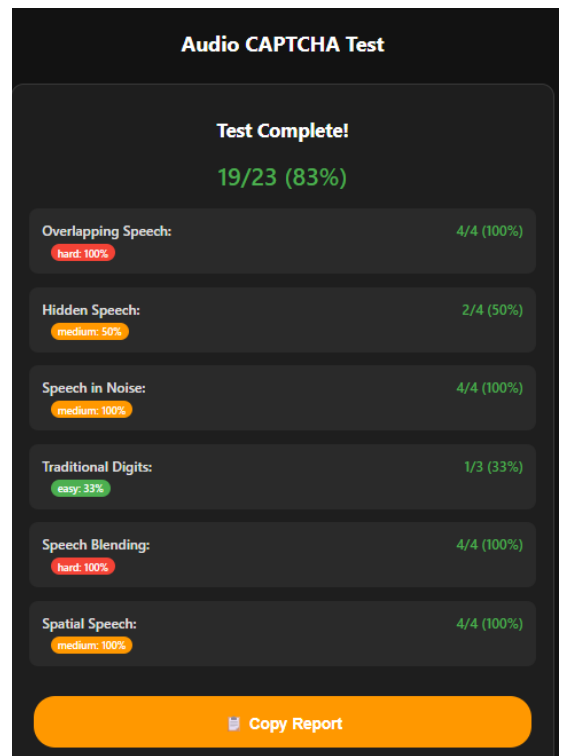

*Figure 4. Human reports on easy-hard captcha challenges*

One possibility is drawing inspiration from the human auditory pathway: incorporating algorithms that perform auditory scene decomposition, separating an audio signal into hypothesized sources or using predictive coding to infer missing pieces. Indeed, in the spatialized and reverberant speech cases, our analysis showed that the model often transcribed the echo or garbled audio literally (e.g. repeating a word because it heard the echo), whereas no human would mistake an echo for a new word. This indicates that current models lack an understanding of the physical environment producing the sound. Approaches that integrate physics-based audio models or that explicitly model reverberation might help align machine perception with human perception in such scenarios. Interestingly, our attempt at an ambiguous audio illusion (where context influences perception) revealed another facet: humans apply top-down cognitive expectations to interpret sound, something AI typically does only if such context is directly encoded. A multimodal system that had access to visual cues or prior textual context might simulate this to some degree, but the pure audio model we tested did not.

This raises the question of how tightly integrated the "language understanding" and "perception" components of an AI need to be to emulate human-like interpretation. Humans seamlessly combine prior knowledge, context, and sensory input – for instance, knowing that a conversation is about birthdays will help a person parse a muffled word as "cake" rather than a similar-sounding alternative.

Current architectures often treat perception (e.g. an audio encoder or a vision encoder) as a separate front-end that feeds into a language model. Our results suggest that a more intertwined approach, where higher-level expectations can dynamically influence lower-level perception (akin to how our brain's auditory cortex is modulated by context), could be crucial for closing the gap. The implications of our auditory Turing test extend beyond just diagnosing model weaknesses – they also chart a path forward for research.

By enumerating concrete tasks where AI fails and humans excel, we provide clear targets for the community to tackle. Some failures may be mitigated by data: for example, training on multi-speaker mixtures or on audio with heavy reverb might improve a model's ability to handle those conditions. But as evidenced by the limited success of brute-force noise training [8], solely increasing training data may not yield human-level performance. Instead, new model architectures or training objectives might be required.

One idea is to incorporate auxiliary tasks that mimic human auditory skills, such as source separation or noise classification, into the training of multimodal models. Another idea is leveraging self-supervised learning on audio in the wild to teach models to handle real-world complexities; just as infants learn to pick out voices by listening in many environments, an AI could use unsupervised signals to develop similar capabilities.

Additionally, techniques from the adversarial domain could be repurposed in a constructive way: by training models on perturbed attack audio that is challenging, we might harden their perception against both random and malicious distortions. It is also worth noting the evaluation perspective our benchmark offers. By treating these perceptual challenges as a "Turing test," we emphasize that the goal is not merely to improve metrics on a narrowly defined task, but to reach human parity on a broad spectrum of real-world listening tasks. This has practical importance. For instance, robust cocktail party listening in AI could revolutionize assistive technologies like hearing aids and voice-controlled interfaces in noisy environments. Similarly, better handling of accented or disfluent speech would make voice assistants more inclusive.

The current inability of AI to handle such cases is a barrier to deploying systems in many human-centric applications. Thus, solving the tasks in our auditory Turing test is not only a benchmark exercise but also a step toward more inclusive and resilient AI. We anticipate that future multimodal LLMs will incorporate specialized auditory front-ends or learned mechanisms that specifically target these capabilities, guided by benchmarks like ours.

## CONCLUSION

We have presented an auditory Turing test framework to evaluate whether AI systems possess the human-like ability to hear and understand complex auditory inputs. By curating a set of audio challenges (overlapping speech, noise-occluded speech, distorted and spatial audio, among others) that humans solve with ease, but today's models largely cannot, we exposed a significant gap in current multimodal AI.

Our results show that even state-of-the-art language models with advanced speech recognition components fail catastrophically on tasks that are trivial for human listeners, underscoring how far we remain from true human-level auditory perception in machines. These findings reaffirm the essence of the cocktail party paradox – that what our ears and brain do effortlessly is still extremely hard for artificial listeners. The contributions of this work are twofold. First, we deliver a diagnostic benchmark that can be used to quantify progress in machine listening on human-favored tasks. Second, through analysis of model failures, we provide insights into why these failures occur and suggest directions for closing the gap. In drawing parallels to recent vision benchmarks that revealed blind spots in AI perception, our study highlights the need to redirect some research attention from purely improving high-level reasoning toward improving low-level perceptual capabilities.

Overcoming the challenges posed in this paper will likely require novel approaches integrating selective attention, robustness to noise, and context-aware perception into the architecture of multimodal systems. We hope that the "cocktail party test" and related tasks proposed here will serve as a litmus test for auditory intelligence, driving the development of AI that can truly listen like a human. Achieving this will not only pass a conceptual milestone (an auditory Turing test) but also unlock more natural and resilient human-computer interaction in everyday noisy, cluttered, and unpredictable environments.

## ACKNOWLEDGEMENT

The authors thank the PeopleTec Technical Fellows program for research support.

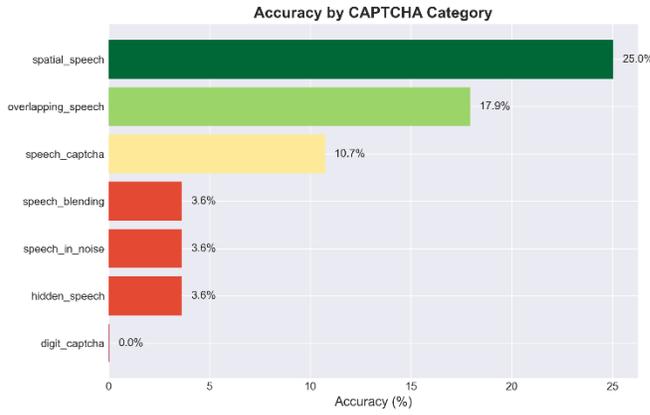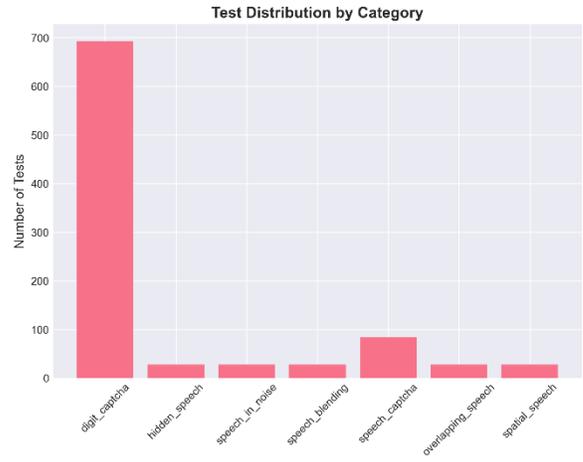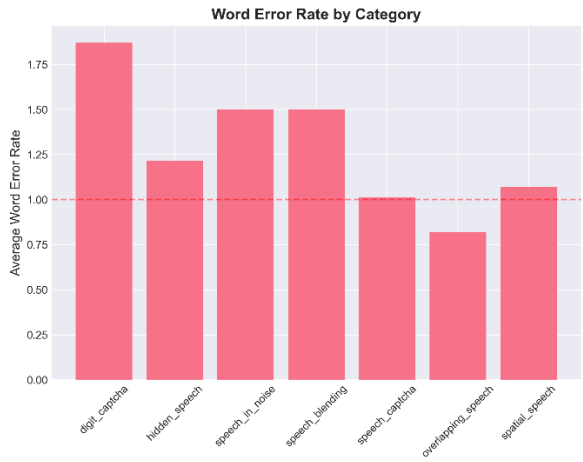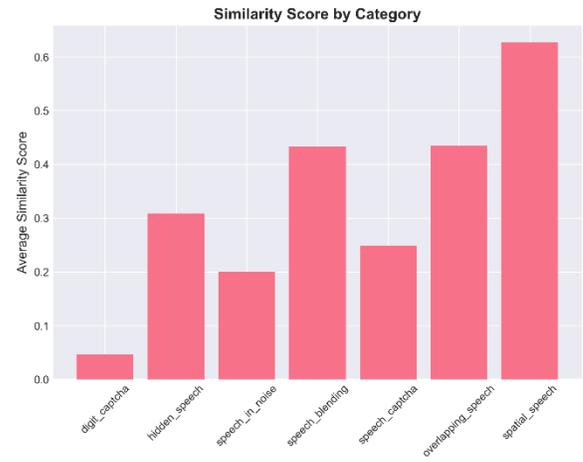

*Figure 5. CAPTCHA Accuracy by Category Across All Models*

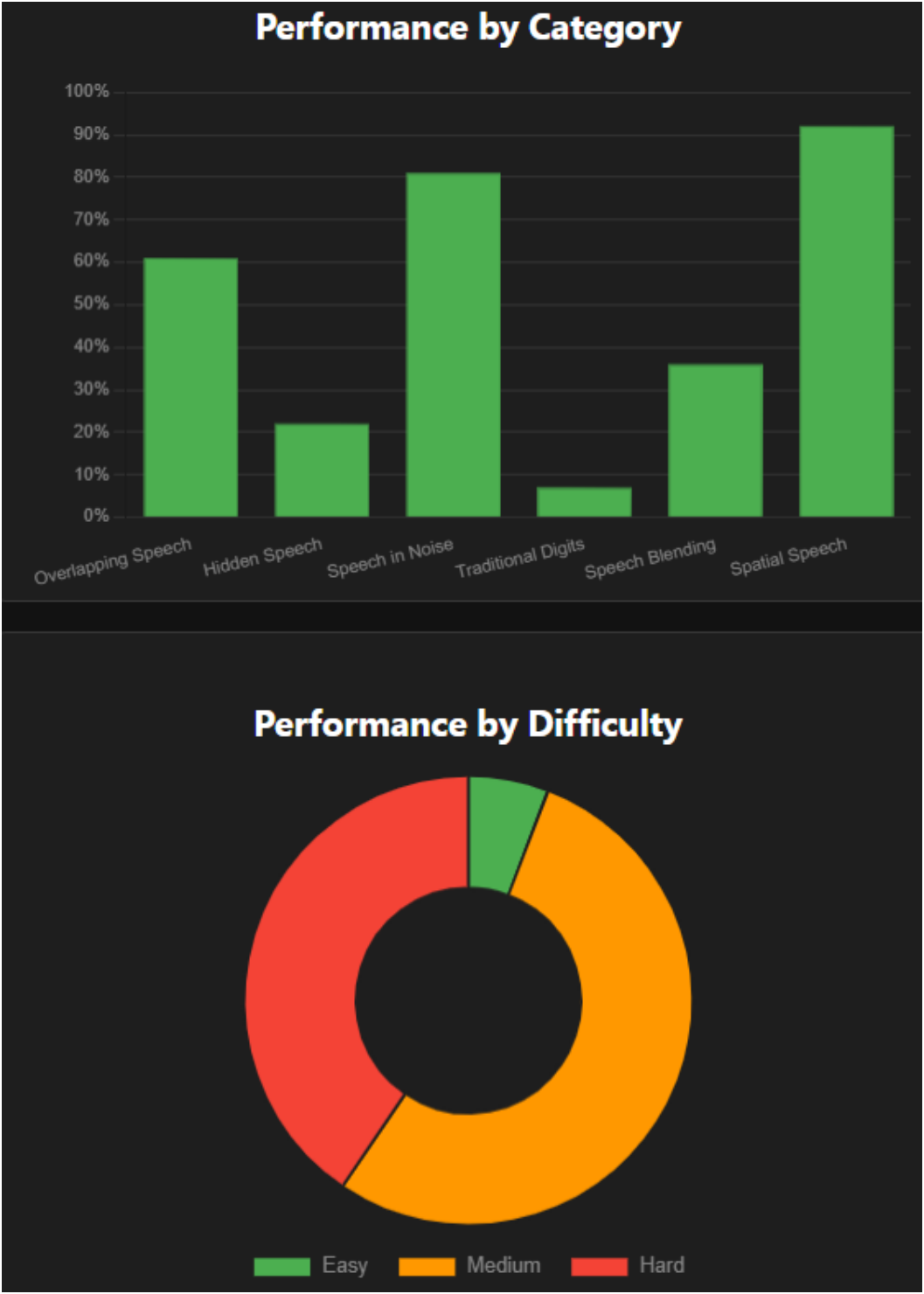

*Figure 6. Summary of Human Test Results by Challenge Categories and Difficulty*

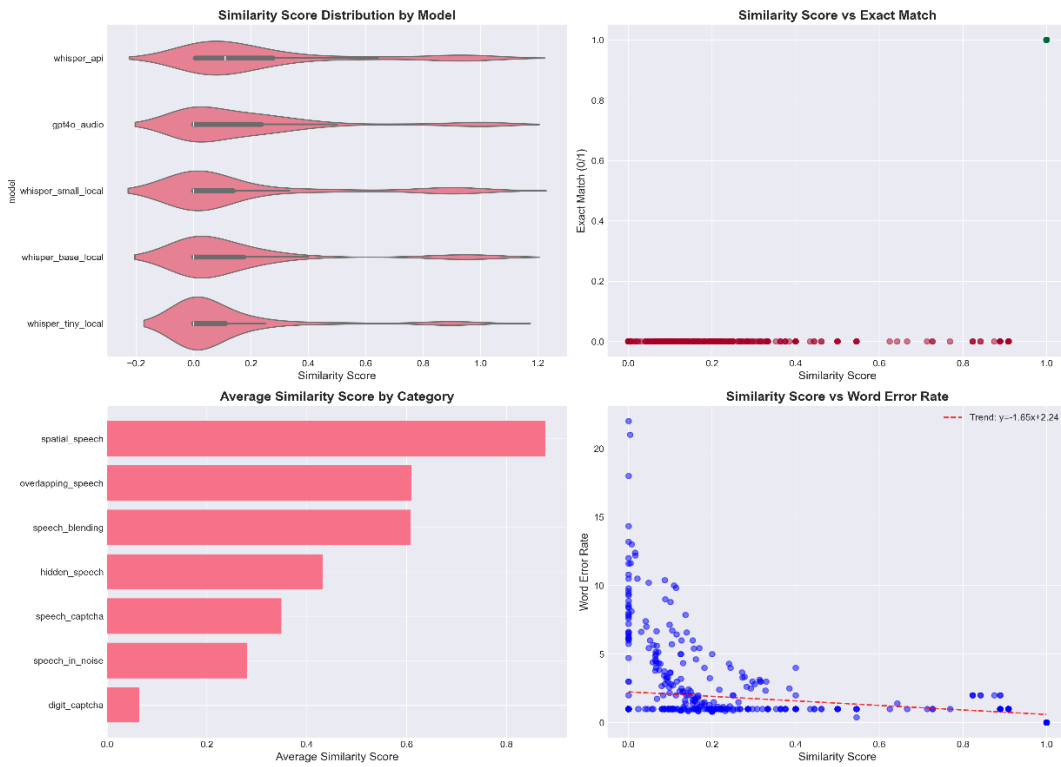

*Figure 7. Similarity Score Analysis*

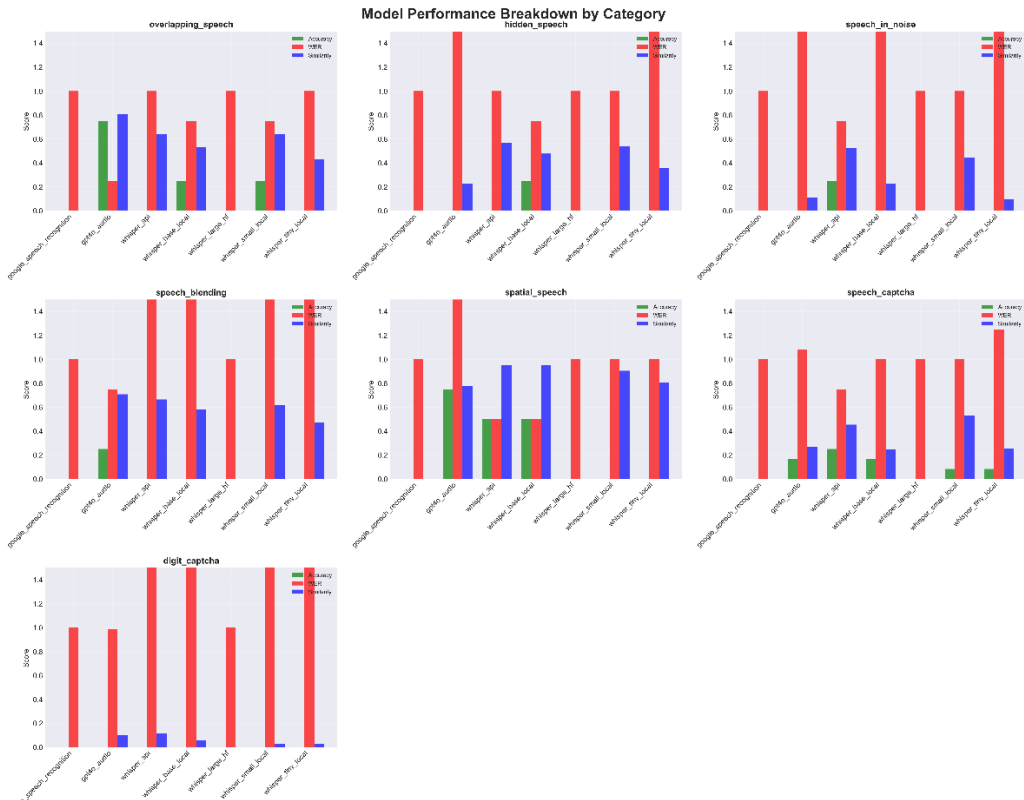

*Figure 8 Model Performance Breakdown by Category*